\documentclass{article}

\usepackage{PRIMEarxiv}

\usepackage{hyperref}
\usepackage{url}
\usepackage{booktabs}
\usepackage{amsfonts}
\usepackage{amsmath}
\usepackage{nicefrac}
\usepackage{microtype}
\usepackage{fancyhdr}
\usepackage{graphicx}
\graphicspath{{fig_/}}
\usepackage{xcolor}
\usepackage{siunitx}
\usepackage{listings}
\definecolor{codegreen}{rgb}{0,0.6,0}
\definecolor{codegray}{rgb}{0.5,0.5,0.5}
\definecolor{codepurple}{rgb}{0.58,0,0.82}
\definecolor{backcolour}{rgb}{0.95,0.95,0.92}

\lstdefinestyle{mystyle}{
    commentstyle= \color{red!50!green!50!blue!50},
    keywordstyle= \color{blue!70},
    numberstyle=\tiny\color{codegray},
    stringstyle=\color{codepurple},
    basicstyle=\ttfamily\footnotesize,
    breakatwhitespace=false,
    breaklines=true,
    captionpos=b,
    keepspaces=true,
    numbers=left,
    numbersep=5pt,
    showspaces=false,
    showstringspaces=false,
    showtabs=false,
    tabsize=2,
    frame=single
}

\title{EvoAgent: An Evolvable Agent Framework with Skill Learning and Multi-Agent Delegation
}
\author{
Aimin Zhang$^{1}$\thanks{Corresponding author: zhangaimin@focuschina.com} \\
$^{1}$Focus AI Center, Focus Technology Co., Ltd. \\
\texttt{zhangaimin@focuschina.com}
\and
Jiajing Guo$^{1}$ \\
$^{1}$Focus AI Center, Focus Technology Co., Ltd. \\
\texttt{guojiajing@focuschina.com}
\and
Fuwei Jia$^{1}$ \\
$^{1}$Focus AI Center, Focus Technology Co., Ltd. \\
\texttt{jiafuwei0921@focuschina.com}
\and
Chen Lv$^{1}$ \\
$^{1}$Focus AI Center, Focus Technology Co., Ltd. \\
\texttt{lvchen1018@focuschina.com}
\and
Boyu Wang$^{2}$ \\
$^{2}$Engineering Research Center of Digital Forensics, Ministry of Education, \\
\ Nanjing University of Information Science and Technology \\
\texttt{wangboyu@focuschina.com}
\and
Fangzheng Li$^{1,3}$ \\
$^{1}$Focus AI Center, Focus Technology Co., Ltd. \\
$^{3}$Nanjing University of Science and Technology \\
\texttt{lifangzheng@focuschina.com}
}
\begin{document}
\maketitle

\begin{abstract}
Large language model (LLM) agents increasingly rely on reusable skills, but skills are usually written by hand, and letting an agent change them on its own calls for explicit user control. We report on EvoAgent, an evolvable agent framework built for a foreign trade assistant, and share the engineering choices behind it. We organize the agent as a harness of four components (reasoning, context engineering, tools and state) and confine self-improvement to the user's persistent state: profile, memory and skills. Skills are multi-file packages with lazily loaded references, matched in three stages (keyword, embedding, LLM). The main agent can delegate to sub-agents with their own context, and long histories are compressed while an index of recoverable resources is kept. An offline loop reviews completed sessions to update the profile and memory and to propose skills; in the open-source release a skill changes only after the user confirms the proposal. On 172 foreign trade test cases scored by an LLM judge on five dimensions, GPT-5.2 with EvoAgent scored about 28\% higher than GPT-5.2 called directly with only the final query. This is a single run judged by the model under test, so it measures the harness as a whole rather than any one mechanism, and swapping in GPT-4.1 or Qwen3.5-35B-A3B lowered the scores relative to calling those models directly. Several mechanisms described here were added after the version we evaluated and are not evaluated. The evaluation data is proprietary, so the results are for reference only. An open-source release of the code, which is not the evaluated business version, is available at \url{https://github.com/Focus-AI-Center/Mentarc-EvoAgent}.
\end{abstract}


\section{Introduction}

Large Language Model (LLM) agents have gradually emerged as a core paradigm for achieving autonomous task execution in complex and open environments. By integrating capabilities such as reasoning, planning, tool invocation, and environment interaction, LLM agents are able to accomplish end-to-end task solving within a unified framework \cite{wang2024survey,yao2022react,schick2023toolformer}. In professional domains such as software engineering, scientific computing, financial analysis, and enterprise workflow orchestration, the tasks faced by agents increasingly exhibit multi-step, cross-domain, and highly structured characteristics. Their complexity often exceeds what isolated calls to atomic tools can support. To bridge this gap, researchers have progressively introduced the concept of “Skill” as a higher-level encapsulation of capability \cite{zhang2026evoskills,claude-skill}. Unlike atomic tools that provide a single function, skills are defined as structured multi-file collections, typically including workflow instructions, executable scripts, domain knowledge references, and reusable logic modules, thereby offering systematic and reusable process support for complex task execution \cite{xu2026agent}.

\par
Although skills help agents with complex tasks, existing ways of acquiring and applying them have several limitations. First, mainstream approaches heavily rely on manual authoring, resulting in high labor costs, limited scalability, and difficulty in maintaining consistent quality across domains. Second, manually constructed skills often exhibit human–AI cognitive misalignment: workflows optimized based on human intuition may not align with the intrinsic reasoning patterns of large language models, leading to performance degradation in highly structured domains such as natural sciences \cite{zhang2026evoskills,li2026skillsbench}. Third, most existing self-evolution methods primarily focus on heuristic optimization at the level of individual tools or prompts, and do not yet support systematic iterative generation, optimization, and validation of complex multi-file skill packages. Finally, many self-improvement frameworks rely on real annotated supervision \cite{zhang2026evoskills} or intensive expert feedback, conditions that are often difficult to satisfy in real-world deployment scenarios.

\par
Meanwhile, advances in multi-agent systems indicate that task decomposition and role specialization can help with complex problems \cite{feng2026multi,tran2025multi}. Within hierarchical orchestration architectures, systems support configurable sub-agents and explicit task delegation: the main agent decomposes high-level objectives and assigns subtasks to specialized sub-agents with domain-specific capabilities. This design effectively mitigates context window limitations, reduces cognitive load, and improves parallel processing capacity and system robustness. However, mainstream multi-agent frameworks typically adopt static role definitions and fixed task routing strategies, lacking adaptive mechanisms that dynamically evolve delegation logic, skill sets, and collaboration patterns based on historical experience. Moreover, the integration between skill self-evolution mechanisms and hierarchical multi-agent task delegation remains limited, resulting in a structural disconnect between individual skill learning and global task scheduling optimization. To address these challenges, we present EvoAgent, an evolvable LLM agent framework that integrates skill learning with hierarchical multi-agent task delegation.

\par

In terms of design, EvoAgent follows harness engineering. The term harness engineering was used by Mitchell Hashimoto in 2026 for the practice of engineering a lasting fix whenever an agent makes a mistake, for example in the agent's instruction file, so that the mistake does not recur \cite{hashimoto2026harness}. OpenAI described a related shift in a report on building a product with Codex agents and no manually written code: engineering work moves from writing code to designing environments, specifying intent and building feedback loops that let agents work autonomously \cite{openai2026harness}.

At the architectural level, Birgitta Böckeler distinguishes guides, which steer an agent before it acts, from sensors, which observe after it acts and help it self-correct, and separates computational controls, which are deterministic and fast, from inferential ones such as AI code review \cite{fowler2026harness}. Furthermore, Paul Iusztin describes the harness as every piece of code, configuration and execution logic that is not the model itself, nests prompt engineering, context engineering and harness engineering as successively broader levels, and frames the LLM as the operating system on which agentic applications are built \cite{iusztin2026agenticharness}.

Unlike skill-based agents that focus on automated co-evolutionary verification, EvoAgent implements harness engineering as a user-centered dual-loop system with online execution and post-session evolution. The online loop constructs a deterministic execution trajectory through enforced skill matching, context assembly, and task delegation pipelines; the offline loop enables capability evolution via asynchronous session review and skill extraction mechanisms. On this basis, EvoAgent aims at controllable capability growth in a foreign trade assistant.
\par
To locate where self-improvement may act, we use a coarse four-component view of a harness: \emph{reasoning} (the backbone model and its ReAct loop), \emph{context engineering} (how the context window is assembled, disclosed and compressed), \emph{tools}, and \emph{state}. State is further divided into \emph{agent state}, which is scoped to a session (dialogue history and compression state), and \emph{user state}, which persists across sessions and belongs to the user (profile, memory and skills). EvoAgent keeps the model weights, the context-engineering logic and the tool implementations fixed and confines self-improvement to the user state: the system serves a user better because the state it reads at the start of each session has been enriched, not because any other component has been rewritten. This is a deliberately bounded form of self-improvement; the improvement procedure itself, that is, the review prompts and update logic, is not modified by the system.
\par
Specifically, the framework includes the following four design choices:

\begin{itemize}
    \item Skills as multi-file packages: reusable procedural knowledge is organized into packages with lazily loaded references and evolutionary metadata, persisted per user and loaded on demand.
    \item User-centered skill evolution: skill usage frequency (and, where a deployment supplies outcome signals, execution success) is tracked, new or revised skills are proposed from reviewed sessions, and no labeled data is required.
    \item Hierarchical sub-agent delegation: the main agent can create specialized sub-agents with independent context spaces on demand, which keeps their working context separate from the main one.
    \item The agent is organized as a harness of four components (reasoning, context engineering, tools and state), self-improvement is confined to a persistent, user-owned state (profile in USER.md, long-term memory in MEMORY.md, and skills), and history compression supports long-term context retention.
\end{itemize}

In summary, EvoAgent is a framework for a foreign trade assistant that combines skill learning with collaborative task execution and confines self-improvement to the user's persistent state. This paper describes its design and the engineering choices behind it, and reports a limited evaluation in that setting (Section~\ref{sec:experiments}).

\section{Related Work}
\label{sec:headings}

\subsection{Skill Learning and Self-Evolution in LLM Agents}

Large language models have made rapid progress in tool invocation and complex task execution \cite{schick2023toolformer,qin2023toolllm}. Among these advances, “Skill” has emerged as a higher-level capability encapsulation distinct from atomic tools, designed to support multi-step, cross-domain professional task solving. Anthropic formally introduced the concept of agent skills \cite{anthropic2025agentskills,claude-skill,xu2026agent}, defining them as structured multi-file packages composed of workflow instructions, executable scripts, and domain reference files, thereby enhancing task completion capabilities in professional scenarios. In this context, constructing agent systems with autonomous planning and tool invocation abilities has become a key research focus. Furthermore, some studies have begun to address the problem of “capability growth,” namely how agents can accumulate experience during continuous task execution and transform it into reusable and transferable capability units. Existing research typically follows several paths \cite{zhang2026evoskills}:

\begin{itemize}
    \item Strategy summarization based on execution trajectories, abstracting decomposition steps of successful tasks into templated operational workflows \cite{yang2026autoskill};
    \item Reflection-based mechanisms that analyze failure trajectories and revise strategies accordingly \cite{qiu2026autorefine};
    \item Retrieval-augmented memory approaches that invoke historical successful cases in new tasks to improve reasoning quality \cite{mugambiwa2026multi,zhou2026mobile}.
\end{itemize}

These systems also differ in where the improvement is written and in what stands between a proposed change and its adoption. SEAgent improves a computer-use agent by updating the model's own policy with reinforcement learning on self-generated tasks \cite{sun2025seagent}, whereas skill-based systems such as Voyager \cite{wang2023voyager}, AutoSkill \cite{yang2026autoskill} and EvoAgent leave the model fixed and write to an external store of skills. Among the latter, AutoRefine admits a new rule, skill or bounded sub-agent only after contract and replay checks show that it fixes the linked failures without regressions \cite{qiu2026autorefine}; EvoAgent places the user's confirmation in this position (Section~\ref{sec:controlled}).

Although these methods have achieved certain improvements in enhancing single-agent stability and task success rates, their core capability structures still rely on fixed prompting frameworks or predefined strategy patterns. To our knowledge, these distinctions are rarely modeled explicitly. Existing systems cover parts of the skill lifecycle, for example generating skills from trajectories and validating them before adoption \cite{qiu2026autorefine}, but an end-to-end treatment of the whole lifecycle (including generation, evaluation, consolidation, and elimination) remains uncommon, as do designs of self-evolving agents targeted at specific scenarios \cite{Hermes}. Moreover, current work mainly focuses on internal strategy optimization within a single agent, with limited exploration of how skill learning mechanisms can co-evolve with system-level architectural expansion.

Therefore, research on evolvable agents still faces two key challenges: first, how to abstract task execution trajectories into composable and transferable structured skill units; second, how to integrate skill learning mechanisms with architectural evolution so that capability expansion is reflected not only at the strategic level but also at the organizational and collaborative levels.

Against this backdrop, EvoAgent treats skills as continuously accumulable structured capability assets, supporting their generation, evaluation, and reuse through a closed-loop feedback mechanism, with the aim of sustained capability growth oriented toward long-term task distributions.

\subsection{Multi-Agent Collaboration and Structured Task Architectures}

In complex task scenarios, large language models often face limitations in context window size and excessive cognitive load. To alleviate these bottlenecks, researchers commonly adopt task decomposition and structured execution mechanisms, breaking down overall objectives into subtasks and enhancing stability and controllability through staged reasoning. A representative method is the ReAct framework, which alternates reasoning and tool invocation to form a “Think–Act–Observe” closed-loop process, thereby strengthening complex task handling capabilities \cite{yao2022react}.

Building upon this, some studies introduce multi-agent collaboration mechanisms, leveraging role specialization and dialog interaction to achieve task decomposition and cooperative problem-solving. For example, CAMEL \cite{li2023camel} simulates collaborative task-solving by assigning different roles to agents engaged in dialogue, while AutoGen \cite{wu2023autogen} proposes an orchestratable multi-agent conversation framework that supports the separation of planning and execution for complex tasks. These approaches typically employ predefined roles and fixed communication protocols to complete tasks through multi-round interactions. Specialist hierarchies follow the same pattern: UltrasoundAgents, for example, lets a main agent localize a lesion and a sub-agent analyze the zoomed region, with the roles defined by the diagnostic pipeline \cite{zhu2026ultrasoundagents}.

However, most existing multi-agent frameworks focus on collaborative reasoning at the level of single tasks. Their collaboration structures are usually statically determined during system design and lack the capability to continuously optimize based on historical interaction outcomes. Recent work does optimize the prompts of a production multi-agent assistant against rubrics and an LLM judge calibrated to human annotations \cite{herrera2026build}, but its optimization target is the agents' prompts. At the same time, role specialization is often primarily organizational in nature and is not deeply integrated with transferable skill accumulation mechanisms, making it difficult to support long-term capability accumulation and system self-evolution.

On the other hand, some works have begun to explore self-improvement and experiential reflection mechanisms in agents. Reflexion \cite{shinn2023reflexion} introduces a language feedback loop to revise strategies; Voyager \cite{wang2023voyager} constructs a skill library to support continuous capability expansion in open environments; Generative Agents \cite{park2023generative} investigate long-term memory and behavioral evolution mechanisms in simulated settings. While these methods emphasize experience accumulation and capability enhancement, their capability updates often remain implicit and lack structured, transferable skill representations.

Overall, existing research has made important progress in task structuring, multi-role collaboration, and reflective capability improvement. However, in real-world business scenarios, there remains a lack of a unified framework that simultaneously integrates structured process control, lightweight multi-role collaboration, and explicit skill accumulation mechanisms. Particularly in industrial environments that emphasize deployment cost and system scalability, achieving sustainable evolution and capability transfer without over-reliance on complex multi-agent orchestration remains an important research challenge.

\subsection{Harness Engineering}

As the capabilities of large language models rapidly improve, research emphasis has gradually shifted from “how to enhance model reasoning ability” to “how to systematically harness model capability.” Harness engineering, a recently emerged paradigm, emphasizes constraining, guiding, and integrating model behavior through an external structured control layer, thereby constructing stable, observable, and evolvable agent systems.

Mitchell Hashimoto named the practice of harness engineering in a 2026 post on adopting AI coding agents: whenever the agent makes a mistake, the engineer builds a lasting fix, for example in the agent's instruction file or through programmed tools, so that the mistake does not recur \cite{hashimoto2026harness}. Effort thus moves from correcting individual outputs to improving the environment in which the agent works.

In the report \textit{Harness engineering: leveraging Codex in an agent-first world}, OpenAI describes building an internal product with Codex agents and no manually written code. It reports that engineering work shifted from writing code to designing environments, specifying intent and building feedback loops and scaffolding for autonomous agents, with practices such as treating documentation as a navigable map, enforcing architectural invariants mechanically, and keeping context inside the repository \cite{openai2026harness}.

Birgitta Böckeler, writing on Martin Fowler's site, distinguishes guides, which steer the agent before it acts, from sensors, which observe after it acts and help it self-correct, and separates computational controls, which are deterministic and fast, from inferential ones such as semantic analysis, AI code review and LLM-as-judge \cite{fowler2026harness}. Crossing the two distinctions gives a structural lens for analyzing how an agent's behavior is regulated.

Anthropic describes a harness for long-running application development in which a planner, a generator and a separate evaluator cooperate and hand work over through structured context resets as an alternative to compaction, and it reports that separating the agent that does the work from the agent that judges it is a strong lever \cite{anthropic2026longrunning}. EvoAgent likewise keeps its offline reviewer separate from the online agent.

Furthermore, Paul Iusztin defines the harness as every piece of code, configuration and execution logic that is not the model itself. He nests three levels: prompt engineering crafts the instructions, context engineering decides what enters the context window and when, and harness engineering covers the full application and infrastructure. A complete harness, in his account, consists of the LLM, tools, a planning loop, context engineering, a sandbox, memory, an orchestration layer and a serving layer, with the LLM acting as the operating system around which applications are built \cite{iusztin2026agenticharness}.

EvoAgent's four-component view (Figure~\ref{fig:architecture}) regroups this list to locate where self-improvement may act. The LLM and the planning loop form \emph{reasoning}, which in EvoAgent is a ReAct cycle; context engineering and tools carry over unchanged; and memory becomes part of \emph{state}, which in EvoAgent also holds the user's profile and skills and the history of the current session. The sandbox, the orchestration layer and the serving layer appear in Figure~\ref{fig:architecture} as the code sandbox, sub-agent delegation and the API entry, but we do not treat them as separate components because the offline loop never modifies them. What distinguishes our view is that state is singled out as the one component that self-improvement writes to.

Compared with traditional research on multi-agent collaboration or skill self-evolution, harness engineering places stronger emphasis on system-level controllability, observability, and evolvability. EvoAgent follows this paradigm and separates the online execution loop from the offline evolution loop.

\section{Methodology}
\label{sec:others}

\begin{figure}
    \centering
    \includegraphics[width=0.85\linewidth]{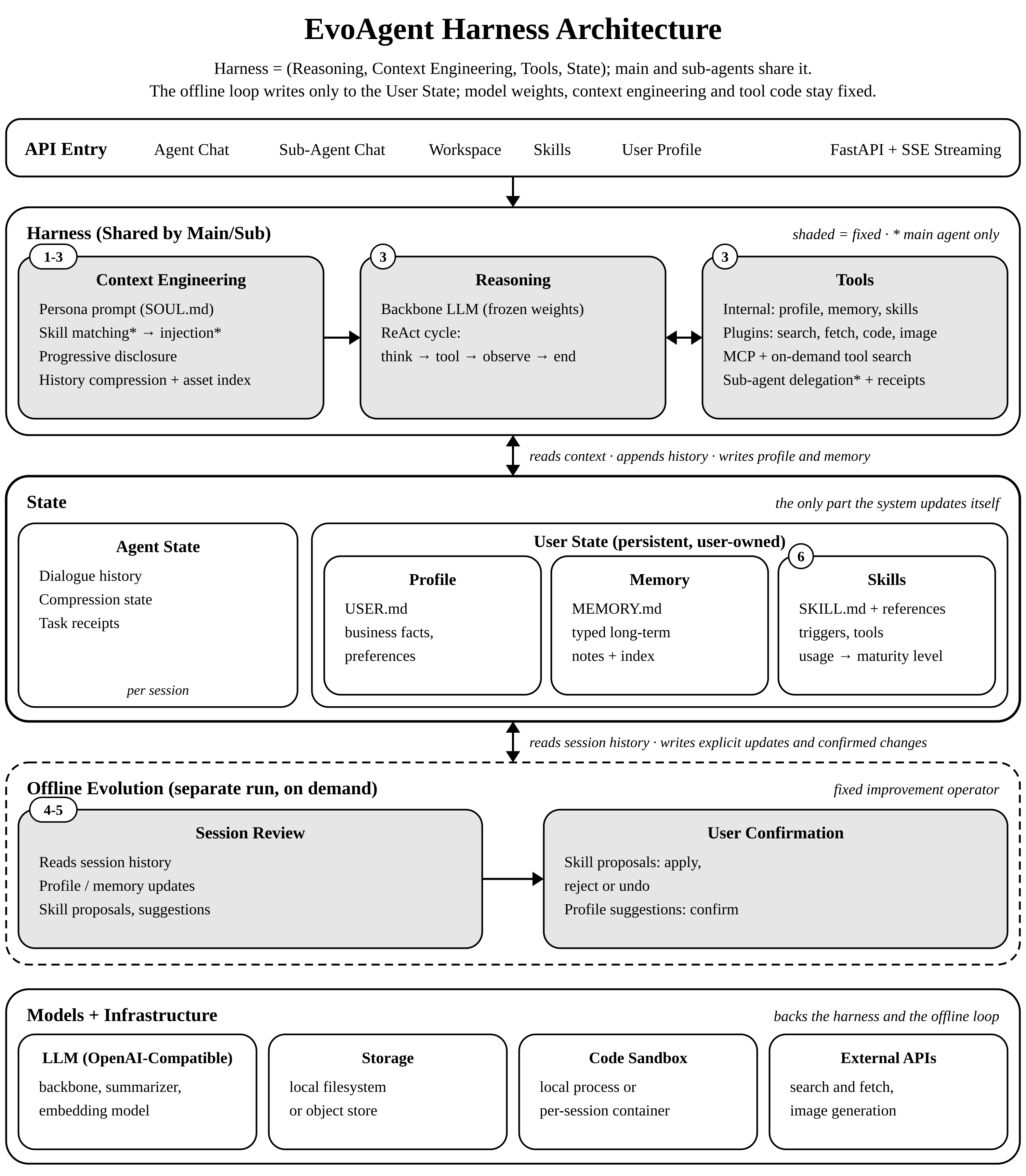}
    \caption{EvoAgent as a harness of reasoning, context engineering, tools and state. Shaded components are fixed during self-improvement; the offline loop (dashed) reads session history and writes only to the user state. Numbers 1--6 mark the execution stages of Section~\ref{3.3}, and * marks functions performed by the main agent only.}
    \label{fig:architecture}
\end{figure}

The EvoAgent framework is a user-centered evolvable agent architecture. Through continued interaction with a user, EvoAgent accumulates and refines skills in that user's persistent state (Section~\ref{sec:controlled}). This section describes the method: an overview, a formal model of EvoAgent, and the system architecture. Figure~\ref{fig:architecture} illustrates the overall architecture of EvoAgent. The EvoAgent architecture adopts a hierarchical design in which a main agent collaborates with sub-agents to accomplish task processing, and it is organized as a harness with four components. \emph{Reasoning} is the backbone LLM driven by a ReAct loop. \emph{Context engineering} assembles what the model sees: the persona prompt (SOUL.md), the user state, the matched skill, and a history compressed into a structured summary with an asset index, with context disclosed progressively. \emph{Tools} comprise internal tools, plugins, MCP servers and sub-agent delegation, and tool schemas can be loaded on demand. \emph{State} is divided into agent state, scoped to one session (dialogue history, compression state and task receipts), and user state, which persists across sessions and belongs to the user (profile in USER.md, long-term memory in MEMORY.md, and skills). The main agent and its sub-agents share this harness; only the main agent performs skill matching, skill injection and delegation. A separate offline evolution loop, started on demand, reads session history and writes only to the user state, so the online loop and the offline loop communicate solely through state. The architectural highlights include a three-stage skill-matching cascade with two delegation paths (skill-triggered handoff and model-initiated delegation), a progressive disclosure strategy for context utilization \cite{anthropic2025agentskills}, and structured outputs delivered as an event stream. Models and infrastructure (LLM APIs, storage, a code sandbox and external APIs) support both loops.

\subsection{Method Overview}

The core objective of EvoAgent is to let an agent working in open foreign trade environments acquire and refine skills from user interaction, without hand-programming each skill or labeled supervision; in the released implementation the user confirms every skill change (Section~\ref{sec:controlled}). Inspired by the traditional \emph{Human-in-the-loop} paradigm \cite{WU2022364}, its evolutionary mechanism can be summarized as a \emph{User-in-the-loop} driven paradigm, where real user interaction feedback serves as the primary signal for capability evolution.

Overall, the EvoAgent workflow consists of the following three stages:

\begin{itemize}
    \item \textbf{Context-aware matching}: Upon receiving user input, EvoAgent performs a three-stage retrieval process--keyword matching, vector embedding similarity matching, and LLM-based semantic matching--to select the skill unit most appropriate to the current task context from the skill repository.
    \item \textbf{Task execution and feedback collection}: The system injects the matched skill into the current context to complete task execution. During this process, it records skill invocation frequency and, where a deployment supplies outcome signals, an execution success estimate; these serve as feedback signals for subsequent skill evolution.
    \item \textbf{Session-based evolutionary update}: After a session, the evolutionary update can be started (on demand in the released implementation, Section~\ref{3.3}). It updates the user profile and long-term memory from the dialogue history and proposes new or revised skills.
\end{itemize}

In Section 3.2, we formally model EvoAgent’s evolutionary mechanism and construct its corresponding Markov Decision Process (MDP) formulation; Section 3.3 further details the overall framework design and modular implementation of EvoAgent.

\subsection{Problem Formulation}
\label{3.2}

\paragraph{Task Environment Definition}
To formally model the self-evolution process of EvoAgent, we represent its task environment as a Markov Decision Process (MDP). We adopt an approximately observable state space to represent the user interaction environment. The task environment of EvoAgent can thus be defined as $\text{M}_{EA} = (S, A, P, R, \gamma)$, where $\text{S}$ denotes the state space, composed of a quadruple including dialogue history, user profile, skill repository state, and context compression state; $\text{A}$ represents the action space of the LLM, including skill selection, tool invocation, message generation, profile and memory updates, skill proposals, and history compression; $\text{P}$ denotes the state transition probability from the current state $\text{S}$ to the next state $\text{S'}$ given action $\text{a}$; the reward function $\text{R(s, a, s')}$ is modeled as a weighted combination of skill maturity, user profile update magnitude, and memory update magnitude; and $\gamma \in [0,1)$ is the discount factor.

\paragraph{State Space Definition}
The state space of EvoAgent is represented as a four-element tuple:
\[
s = (h, u, S_{\text{skills}}, c).
\]

The tuple splits into agent state $(h, c)$, which is scoped to a session, and user state $(u, S_{\text{skills}})$, which persists across sessions and belongs to the user; only the user state is written by the offline loop (Section~\ref{3.3}). The meanings of each component are summarized in Table~\ref{tab:state_space_components}:

\begin{table}[htbp]
  \centering
  \caption{State Space Components}
  \label{tab:state_space_components}
  \begin{tabular}{lll}
    \toprule
    \textbf{Component} & \textbf{Meaning} & \textbf{Corresponding Content} \\
    \midrule
    \multicolumn{3}{l}{\emph{Agent state (per session)}} \\
    $\mathbf{h}$ & Dialogue history & Session records \\
    $\mathbf{c}$ & Compression state & History compression level \\
    \midrule
    \multicolumn{3}{l}{\emph{User state (persistent, user-owned)}} \\
    $\mathbf{u}$ & User profile and memory & USER.md + MEMORY.md \\
    $\mathbf{S_{skills}}$ & Skill repository state & configs/skills/\{name\}/ \\
    \bottomrule
  \end{tabular}
\end{table}

\paragraph{Skill Space Definition}
A skill $\text{skill} \in S_{skills}$ is formally defined as:
\[
\text{skill = (name, desc, triggers, instr, refs, meta)}.
\]
Here, $\text{name}$ denotes the skill name; $\text{desc}$ denotes the skill description; $\text{triggers}$ is a list of keyword triggers used for matching; $\text{instr}$ is the Markdown-formatted execution instruction; $\text{refs}$ is a dictionary of reference files supporting lazy loading; and $\text{meta}$ represents evolutionary metadata structured as $\text{meta}=(\text{usage\_count}, \text{success\_rate}, \text{created\_at}, \text{updated\_at})$, corresponding to usage count, success rate, creation time, and last update time.

\paragraph{Skill Matching}
EvoAgent employs a three-stage skill matching strategy to balance efficiency and accuracy. The overall matching function is defined as:
\begin{equation}
    M(u, S_{skills}) \rightarrow \text{skill} \cup \{\emptyset\}
\end{equation}

\textbf{Stage 1: Keyword Matching}\\
In this stage, the system traverses all trigger lists of available skills, performing case-insensitive exact string matching, and returns the first matched result.
\begin{equation}
    M_{\text{keyword}}(u, S_{\text{skills}}) = \text{first}\{ \text{skill} \in S_{\text{skills}} \mid \exists t \in \text{skill.triggers},\ \text{lower}(t) \sqsubseteq \text{lower}(u) \}
\end{equation}
where $\sqsubseteq$ denotes substring containment and $\text{first}\{\cdot\}$ returns the first match in a fixed skill order, or $\emptyset$ if there is none.

\textbf{Stage 2: Embedding Matching}\\
Executed only if keyword matching fails, this stage encodes the user input using an external embedding API and computes cosine similarity with existing skill descriptions. The skill with the highest similarity exceeding a predefined threshold is selected. Here, $\text{E}(\cdot)$ denotes the embedding function, $\text{cos}(\cdot)$ denotes cosine similarity, and $\theta$ is the match threshold (0.75 in our experiments).
\begin{equation}
    M_{\text{embed}}(u, S_{\text{skills}}) =
    \begin{cases}
        \text{skill}^{*}, & \text{if } \cos\big(E(u), E(\text{skill}^{*}.\text{desc})\big) \ge \theta, \\
        \emptyset, & \text{otherwise,}
    \end{cases}
    \qquad \text{skill}^{*} = \operatorname*{argmax}_{\text{skill} \in S_{\text{skills}}} \cos\big(E(u), E(\text{skill.desc})\big)
\end{equation}

\textbf{Stage 3: LLM Matching}\\
If the first two stages fail, the system invokes the LLM for intent classification:
\begin{equation}
    M_{\text{llm}}(u, S_{\text{skills}}) = \text{LLM}_{\text{intent\_classify}}(u, S_{\text{skills}})
\end{equation}
The overall function $M$ returns the result of the first stage that does not return $\emptyset$:
\begin{equation}
    M(u, S_{\text{skills}}) =
    \begin{cases}
        M_{\text{keyword}}(u, S_{\text{skills}}), & \text{if } M_{\text{keyword}}(u, S_{\text{skills}}) \neq \emptyset, \\
        M_{\text{embed}}(u, S_{\text{skills}}), & \text{else if } M_{\text{embed}}(u, S_{\text{skills}}) \neq \emptyset, \\
        M_{\text{llm}}(u, S_{\text{skills}}), & \text{otherwise.}
    \end{cases}
\end{equation}

\paragraph{Full State Transition}
Based on the above definitions, a complete state transition of EvoAgent is defined as follows:
\[
\tilde{\text{h}}^{(t+1)} = \text{append}(\text{h}^{(t)}, a^{(t)}, \text{result}^{(t)}),
\]
\[
(\text{h}^{(t+1)}, c^{(t+1)}) = \text{compress}(\tilde{\text{h}}^{(t+1)}, c^{(t)}),
\]
\[
\text{u}^{(t+1)} = f_u(\text{u}^{(t)}, \tilde{\text{h}}^{(t+1)}),
\]
\[
S_{\text{skills}}^{(t+1)} = f_s(S_{\text{skills}}^{(t)}, \tilde{\text{h}}^{(t+1)}),
\]
\begin{equation}
    s^{(t+1)} = (\text{h}^{(t+1)}, \text{u}^{(t+1)}, S_{\text{skills}}^{(t+1)}, c^{(t+1)})
\end{equation}
Here $\text{compress}$ replaces the history by a structured summary and an asset index when its length reaches a threshold $\lambda$, updating the compression state $c$, and otherwise returns it unchanged. The functions $f_u$ and $f_s$ act as the identity except when the user state is actually updated, that is, by a profile or memory tool call or by a confirmed skill proposal (Section~\ref{sec:controlled}); the dialogue history $h$ grows at every step, whereas $c$ changes only when compression runs. The transition is deterministic given the action $a^{(t)}$ and the result $\text{result}^{(t)}$; the randomness of $P$ lies in the action chosen by the LLM and in the result returned by tools and the user.

\paragraph{Reward Function}
In practical EvoAgent applications, there is no explicit numerical reward calculation. Skill tracking primarily relies on implicit monitoring of usage count and, where available, success rate. For theoretical formulation, we define the reward function as a weighted combination of skill maturity, user profile update magnitude, and memory update magnitude, assuming equal weights ($w_1 = w_2 = w_3$). Here, $\Delta u_{\text{profile}}$ denotes the magnitude of user profile updates, and $\Delta \text{memory}$ denotes the magnitude of memory updates:
\begin{equation}
    R(s, a, s') = w_1 \cdot \text{Maturity}(\text{skill}) + w_2 \cdot \Delta u_{\text{profile}} + w_3 \cdot \Delta \text{memory}
\end{equation}

\paragraph{Optimization Objective}
The optimization objective of EvoAgent is to maximize the expected cumulative reward over a time horizon $T$, meaning that from the user’s perspective, EvoAgent is intended to become more effective through continued use, an effect that we do not measure (Section~\ref{sec:threats}):
\begin{equation}
\label{EQ7}
    \max_{\pi} \mathbb{E}_{\pi} \left[ \sum_{t=1}^{T} \gamma^t \cdot R(s^{(t)}, a^{(t)}, s^{(t+1)}) \right]
\end{equation}
where $\pi$ is the policy that selects actions in $A$.

The expected cumulative reward maximization objective defined in Equation~\ref{EQ7} is not computed through explicit numerical reward signals in engineering practice. Instead, it is achieved through a series of \textbf{implicit optimization} mechanisms described in Section~3.3. Specifically:

\begin{itemize}
    \item The update magnitudes $\Delta u_{\text{profile}}$ and $\Delta \text{memory}$ are intended to reflect a deeper system understanding of the user and to improve contextual relevance in subsequent interactions;
    \item Compiling the MDP optimization objective into an asynchronous offline evolutionary loop is a key design choice that lets the evolution loop run in the absence of explicit reward signals.
\end{itemize}

\subsection{EvoAgent Framework}
\label{3.3}

EvoAgent implements the MDP model of Section~3.2 as a six-stage execution process. The core idea is to clearly divide system behavior into two coordinated closed loops:

\textbf{Online Execution Loop}: Handles user requests in real time. This loop favors low latency and fixed, deterministic steps.

\textbf{Offline Evolution Loop}: Runs asynchronously between sessions, learning from historical interactions and optimizing long-term system capabilities. This loop prioritizes deep analysis and systemic optimization.

The coordination of these two loops enables EvoAgent to maintain smooth real-time interaction while continuously enhancing its capabilities. As illustrated in Figure~\ref{fig:architecture}, the overall execution process is divided into the following six stages:

\paragraph{(1) Three-Stage Skill Matching (Online Execution Loop)}
As the entry point of task execution, the system employs a cascaded matching strategy as described in Section~3.2. It first performs low-cost heuristic screening via trigger words; if unsuccessful, it upgrades to embedding-based semantic similarity matching; finally, it resorts to LLM-based intent understanding. This order tries the cheapest method first and calls the LLM only when the other two fail; we did not evaluate matching accuracy.

\paragraph{(2) Skill Injection and Context Assembly (Online Execution Loop)}
Once a skill is matched, the system performs structured context assembly rather than simply appending text to the prompt. This includes injecting the main instruction file \texttt{SKILL.md} and necessary \texttt{references/} files on demand, while updating \texttt{usage\_count} in metadata. The assembled context combines the fixed persona prompt (SOUL.md) with the user state: the profile (USER.md), the memory index and the skill catalog. This stage ensures the agent is equipped with the required domain knowledge before execution.

\paragraph{(3) Task Execution and Dynamic Compression (Online Execution Loop)}
The agent executes the task based on the assembled context. To address context window pressure caused by long conversations, the system monitors token usage in real time and triggers a history compression module when approaching the limit. Unlike conventional summarization, EvoAgent prioritizes preserving structured information (e.g., skill references, external links, key data) and forms an “asset index” to maintain reasoning integrity.

\paragraph{(4) Session Termination Detection and Evolution Trigger (Offline Loop Entry)}
The offline loop runs outside the online request path. In the released implementation it is started on demand, either through the memory-update action or by delegating to the \texttt{session-reviewer} sub-agent. This context-switching point ensures that computation-intensive evolutionary tasks do not interfere with real-time services.

\paragraph{(5) Asynchronous Evolution Update (Offline Evolution Core)}

For user profile and memory management, EvoAgent establishes a multidimensional and dynamically updated information maintenance system structured around five mechanisms:

\begin{itemize}
    \item \textbf{Real-time collection mechanism}: Immediately updates the profile based on user-provided business information during dialogue;
    \item \textbf{Initial guidance mechanism}: Proactively collects key dimensions such as main products and target markets during first interaction;
    \item \textbf{Post-session analysis mechanism}: Extracts new information after sessions to update profile and memory;
    \item \textbf{Behavior suggestion mechanism}: Generates profile update suggestions based on behavioral analysis, finalized upon user confirmation;
    \item \textbf{Response guidance mechanism}: Appends 1–2 guiding suggestions after each task to gradually refine user profile information.
\end{itemize}

This strategy strictly adheres to safety boundary principles, updating profiles only based on explicitly provided user information and avoiding inference of implicit business data from task content.

\paragraph{(6) Skill Maturity Evaluation (Derived from Skill Metadata)}
Based on metadata such as \texttt{usage\_count} and \texttt{success\_rate}, skills are categorized into four levels: Emerging, Growing, Established, and Expert. This evaluation gives users an intuitive indication of how established a skill is. In the released implementation, \texttt{success\_rate} is initialized to 1.0 and is not updated automatically: estimating it requires outcome signals from a deployment's business workflow, which are proprietary and not part of the open-source release, so maturity there reduces to the usage thresholds.

Through this online–offline design, the online loop handles requests with fixed, deterministic steps, while the offline loop is intended to let the user state improve over time. Detailed pseudocode of the evolutionary process is provided in Appendix~A.

\subsection{Scope and Control of Self-Improvement}
\label{sec:controlled}

\paragraph{What may change.} In the four-component view, self-improvement is confined to the user state; the model weights, the context-engineering logic and the tool implementations are not modified by the system. The three parts of the user state follow different write paths. The profile is updated directly only from information the user explicitly provides, either by the agent's profile tool during dialogue or by the session reviewer; a change that goes beyond explicit statements, such as a correction suggested by evidence repeated across conversations, is shown to the user as a suggestion and applied only after the user confirms. Long-term memory is written by the memory tool, on the user's request or by the session reviewer. Skills are never written directly by an agent.

\paragraph{Controlled skill changes.} In the released implementation, the \texttt{skill} tool offers read actions and \texttt{propose}. A proposal (create, update, or an exact-fragment patch) is validated, stored on the server, and shown to the user together with every file it would write; it is applied only when the user confirms, and confirmation refers to the stored proposal, so what is applied is what the user saw. An update is refused if the skill changed after the proposal was made, reference sizes are bounded, and one previous version is retained so that a change or a deletion can be undone. In the earlier open-source release (v1.0), confirmation of an extracted skill was enforced only by the front end (Appendix~C).

\paragraph{Verification.} The proposal workflow of the released implementation is covered by 55 unit tests (\texttt{tests/unit/test\_skill\_proposals.py}), all of which pass on the released revision. They check that the skill tool exposes only read actions and \texttt{propose}, that a staged proposal is not a skill until it is confirmed, and that the model receives only a receipt while the user interface receives the full proposal; that oversized, duplicated or path-escaping references and unsuitable create, update or patch requests are rejected without staging anything; that an update is refused when the skill changed after it was proposed, including a concurrent write during confirmation, while ordinary usage counting does not make a proposal stale; that confirmation can be retried after a partial failure and is harmless when repeated; that proposals expire, are capped per user and are superseded according to fixed rules; that a confirmation applies the staged copy rather than the request, and that a proposal identifier cannot address other files; and that undo restores the previous text and references, including after a patch or a deletion. These tests check the mechanism, not its effect on task performance, which we do not evaluate.

\paragraph{Boundaries of the user state.} Skills are not inert text. A skill can name registered plugins, which are mounted when the skill is loaded, and it carries reference files that the agent reads on demand. A skill therefore shapes which existing tools the agent has at hand, although it cannot add tool implementations or change the model. This coupling between state and tools makes explicit user control over skill changes important.

\paragraph{What is not claimed.} The improvement procedure itself, that is, the review prompts, the extraction rules and the update logic, is fixed: the system improves the state it reads, not the procedure that produces the improvements, so we do not claim recursive self-improvement. Our experiments (Section~\ref{sec:experiments}) compare EvoAgent-enhanced and bare models on a fixed set of test queries; they do not measure how performance changes as state accumulates over sessions. The mechanisms in this subsection describe the released implementation and are not evaluated here.

\section{Experiments}
\label{sec:experiments}

All experiments in this section were run on the stable business version of EvoAgent that was deployed at the time (the \emph{evaluated version} below), with an embedding match threshold of 0.75. The results are reported as originally measured, in a single run on that version; because the business data cannot be released, they are for reference only (see Section~\ref{sec:threats}).

This section presents experiments centered around three Research Questions (RQs): (1) Does EvoAgent as a whole improve the responses of its backbone model in foreign trade scenarios? (2) Is EvoAgent model-agnostic in terms of capability transferability, i.e., how does its overall performance change when the underlying model is replaced? (3) What are the capability boundaries of EvoAgent?

\subsection{RQ1}
\label{4.1:RQ1}

In this experiment, GPT-5.2 is used as the base model of EvoAgent. A ReAct-based framework is designed to support complex workflows in foreign trade scenarios. To evaluate the effectiveness of the EvoAgent system as a whole, we compare two settings: (i) direct invocation of GPT-5.2, and (ii) GPT-5.2 integrated with the EvoAgent framework.

\subsubsection{Test Case Collection}
\label{4.1.1}

To ensure the validity and professionalism of the test data in foreign trade scenarios, we interact with a deployed stable version of EvoAgent via scripted procedures, collecting 664 high-quality multi-turn dialogue sessions. Each session contains 8–9 interaction turns, covering typical foreign trade business scenarios and professional queries. From these sessions we distill 172 evaluation instances, each consisting of the preceding turns as context and a final query to be answered. This dataset is also used in Section~\ref{4.2:RQ2} to ensure consistency across experiments.

It is important to clarify that the expert-level foreign trade questions in the dataset are not manually selected from successful skill-triggering cases. Instead, user queries are programmatically generated via a parameterized script that exhaustively traverses combinations of product categories, target markets, buyer types, and 12 predefined foreign trade scenarios (e.g., market analysis, quotation response, customs clearance and payment risk).

Each query is constructed independently of EvoAgent’s runtime skill matching results, and is not conditioned on whether a skill is successfully triggered or executed. Therefore, the resulting dataset represents a structured sampling of the foreign trade problem space rather than a biased selection favoring EvoAgent.

Although these queries are synthetically generated, data collection is conducted within a deployed EvoAgent environment. Therefore, the distribution is consistent with the deployment setting and is not fully independent of the system. Future work will further evaluate the framework using independently collected foreign trade corpora to test cross-distribution generalization.

\subsubsection{EvoAgent as a Capability Amplifier}
\label{4.1.2}

\paragraph{Capability Enhancement of GPT-5.2 via EvoAgent}

Based on the test samples constructed in Section~\ref{4.1.1}, we extract professional queries and feed them into (i) standalone GPT-5.2 and (ii) GPT-5.2 integrated with EvoAgent, respectively. In the scripts used for these experiments the standalone model receives only the final query of each instance, whereas EvoAgent runs its full loop on the same query (see Section~\ref{sec:threats}). The outputs are then evaluated across multiple dimensions.

\paragraph{Evaluation Design for Model Responses}

We evaluate responses with an LLM-as-Judge approach tailored to our scenario \cite{zheng2023judging} (see Table~\ref{tab:evaluation_dimensions}), scoring them across five dimensions: professionalism, accuracy, completeness, practicality, and language quality. Each dimension is scored on a 1–5 Likert scale, covering terminology correctness, factual accuracy, coverage completeness, actionable feasibility, and linguistic fluency.

To reduce position bias, we randomly shuffle the order of candidate responses during evaluation. In those scripts, each pair of responses is judged in a single call at temperature 0, and the per-dimension means are taken over the pairs that could be scored. Final scores are computed by aggregating the five dimensions, followed by grouping analysis based on intent type and task difficulty.

\begin{table}[htbp]
  \centering
  \caption{Evaluation Dimensions and Descriptions}
  \label{tab:evaluation_dimensions}
  \begin{tabular}{lll}
    \toprule
    \textbf{Dimension} & \textbf{Label} & \textbf{Description} \\
    \midrule
    Professionalism & professionalism & Correctness of domain terminology and logic \\
    Accuracy & accuracy & Factual/policy/data correctness \\
    Completeness & completeness & Coverage of all key aspects of the query \\
    Practicality & practicality & Actionability of the response \\
    Language Quality & language\_quality & Fluency, format, and readability \\
    \bottomrule
  \end{tabular}
\end{table}

\paragraph{Results Comparison}

We compare GPT-5.2 with and without EvoAgent integration. Results are shown in Table~\ref{tab:gpt_comparison}.

In the professionalism dimension, EvoAgent-enhanced GPT achieves 4.762, substantially higher than the baseline (2.703), with an improvement of 2.059. In accuracy, EvoAgent achieves 4.238 compared to 2.907, improving by 1.331.

For completeness, the difference is marginal (4.215 vs. 4.331, -0.116), indicating stability. In practicality and language quality, EvoAgent also shows consistent improvements, reaching 4.709 and 4.779 respectively.

Overall, the average score increases from 3.547 (GPT-5.2) to 4.541 (EvoAgent + GPT-5.2), representing an improvement of about 28\%. These results indicate that EvoAgent substantially improves judged model performance in professionalism, accuracy, practicality, and language quality, while maintaining stability in completeness.

\begin{table}[htbp]
  \centering
  \caption{Comparison of GPT and EvoAgent-Enhanced GPT Across Five Dimensions}
  \label{tab:gpt_comparison}
  \begin{tabular}{l S[table-format=1.3] S[table-format=1.3] S[table-format=1.3]}
    \toprule
    \textbf{Dimension} & {\textbf{GPT}} & {\textbf{EvoAgent GPT}} & {\textbf{Difference}} \\
    \midrule
    professionalism & 2.703 & 4.762 & 2.059 \\
    accuracy        & 2.907 & 4.238 & 1.331 \\
    completeness    & 4.331 & 4.215 & -0.116 \\
    practicality    & 3.744 & 4.709 & 0.965 \\
    language\_quality & 4.052 & 4.779 & 0.727 \\
    \midrule
    \textit{Total / Avg} & 3.547 & 4.541 & {28.0\%} \\
    \bottomrule
  \end{tabular}
\end{table}

\subsection{RQ2}
\label{4.2:RQ2}

\paragraph{Model Selection}

Unlike research-oriented evolutionary agents, EvoAgent is deployed in a real-world foreign trade assistant system. Therefore, in evaluating model transferability, inference cost and deployment feasibility are also key considerations. GPT-5.2 relies on API calls, which can be costly at scale. To examine whether other backbones can be used with EvoAgent, we conduct model substitution experiments.

We select GPT-4.1 \cite{openai2025gpt41} and Qwen3.5-35B-A3B \cite{qwen2026} as baselines, representing closed-source API-based and open-source local deployment paradigms respectively.

\paragraph{Results}

Evaluation metrics are consistent with Section~\ref{4.1:RQ1}. We focus on (1) performance change after EvoAgent integration, and (2) performance gap relative to GPT-5.2.

\subparagraph{GPT-4.1 Analysis}

After integration with EvoAgent, GPT-4.1 shows an average performance drop of approximately 13\% across five dimensions. Compared to EvoAgent-enhanced GPT-5.2, its overall performance is about 75\%.

This suggests that EvoAgent does not improve GPT-4.1 under the current configuration, and even introduces performance degradation. This may be due not only to model capability differences, but also to prompt-style mismatch and instruction-following stability issues. Thus, results reflect the interaction between model and system implementation rather than pure architectural gain.

\subparagraph{Qwen3.5-35B-A3B Analysis}

For Qwen, results show that:

(1) After EvoAgent integration, performance drops to about 85\% of its standalone version;
(2) Compared to EvoAgent-enhanced GPT-5.2, Qwen achieves 74.5\%.

\begin{figure}
    \centering
    \includegraphics[width=0.75\linewidth]{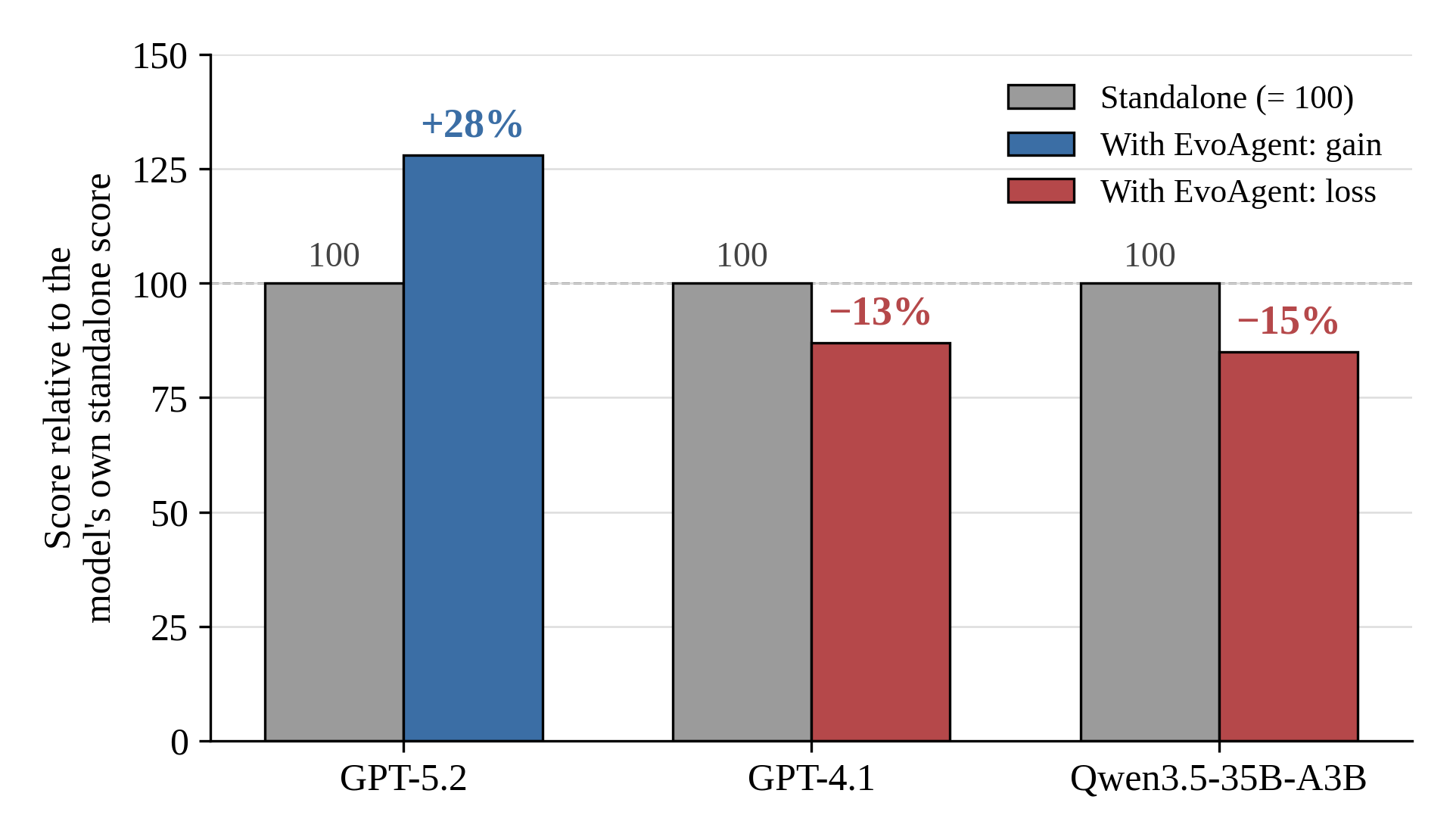}
    \caption{Performance before and after EvoAgent integration, relative to each model's own standalone score (= 100). GPT-5.2 improves by about 28\% (Table~\ref{tab:gpt_comparison}); GPT-4.1 and Qwen3.5-35B-A3B decrease by about 13\% and 15\%.}
    \label{fig:integration_impact}
\end{figure}

\begin{figure}
    \centering
    \includegraphics[width=0.75\linewidth]{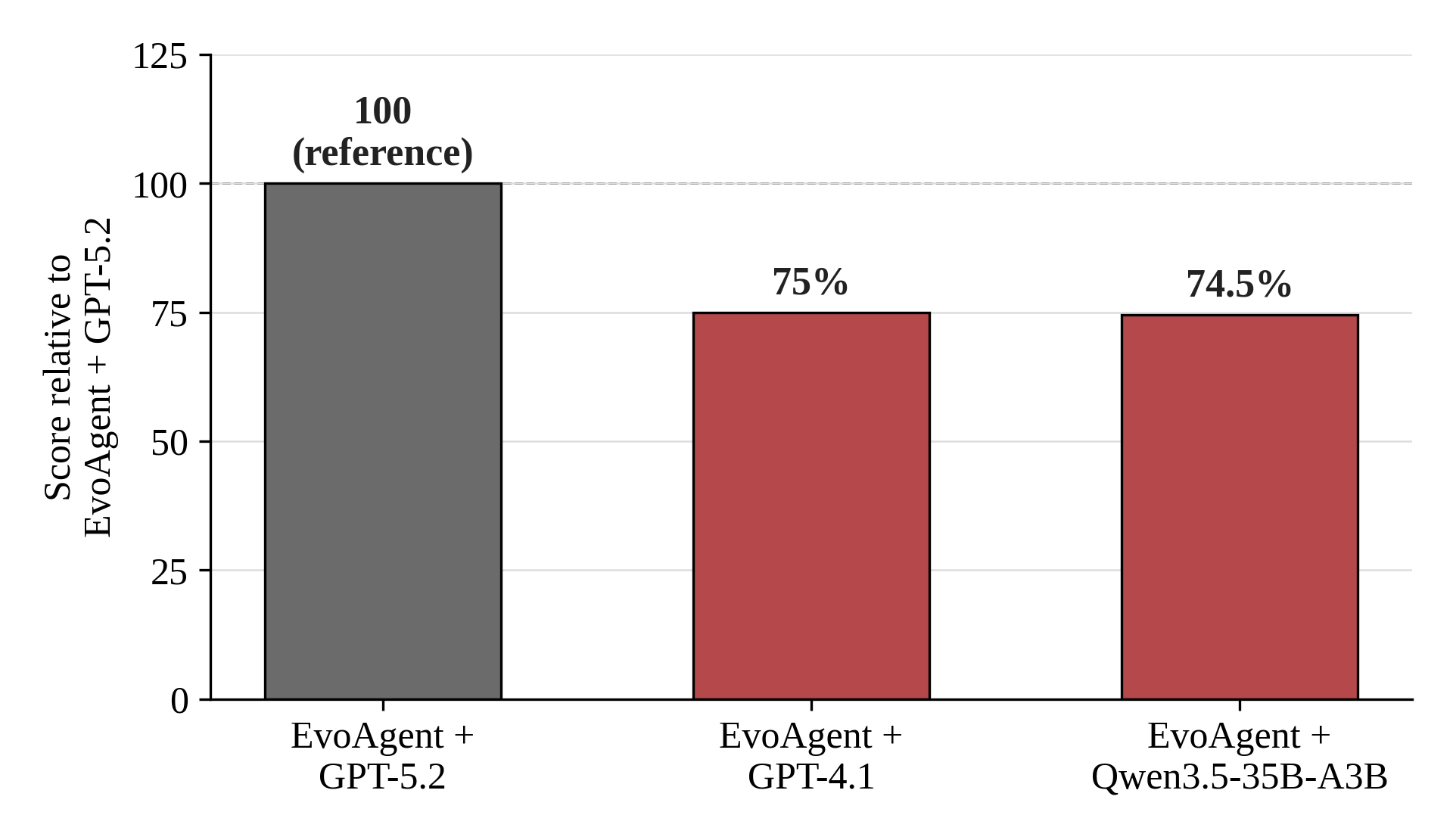}
    \caption{GPT-4.1 and Qwen3.5-35B-A3B with EvoAgent, relative to EvoAgent-enhanced GPT-5.2 (= 100). The values are relative changes only; absolute scores for these models are not reported.}
    \label{fig:relative_comparison}
\end{figure}

Figure~\ref{fig:integration_impact} shows the change of each model relative to its own standalone score, and Figure~\ref{fig:relative_comparison} shows GPT-4.1 and Qwen with EvoAgent relative to EvoAgent-enhanced GPT-5.2. Both figures are normalized and show relative changes only: we report no absolute scores for GPT-4.1 and Qwen, and we did not measure inference cost, so we draw no conclusion about cost-performance trade-offs. Because the judge in those scripts followed the backbone under test (Section~\ref{sec:threats}), the comparison across backbones in Figure~\ref{fig:relative_comparison} is indicative only.

\paragraph{Impact of EvoAgent on Model Capability}

Overall, EvoAgent exhibits heterogeneous effects across models. For GPT-5.2, its structured workflow and multi-stage constraints align well with strong reasoning and tool-use capabilities, leading to performance gains.

In contrast, GPT-4.1 and Qwen show performance instability under structured multi-stage constraints. This may be jointly influenced by model capability, prompt compatibility, skill injection format, and context compression strategies. Without ablation studies, we cannot isolate the contribution of each factor.

Thus, we interpret model performance as an interaction between intrinsic reasoning capability and system-level compatibility:

\begin{equation}
\label{model_cap_fun_}
    \text{Model\_capability} = \text{Model\_Inference\_Capability} \times \text{Agent\_Implementation\_Compatibility}
\end{equation}
In this form, the compatibility term acts as a factor on the model's standalone capability: a value above one means that the agent system amplifies the model, as observed for GPT-5.2, and a value below one means that it reduces it, as observed for GPT-4.1 and Qwen. We use this form only as an interpretive device; neither term is estimated.

\subsection{RQ3}
\label{4.3:RQ3}

\subsubsection{Capability Boundary}

\paragraph{What EvoAgent Can Do}

EvoAgent is deployed as a foreign trade assistant with three main functions.

First, \textbf{product reconstruction}. Given competitor links, product images, or related materials, the system can automatically analyze product structure and generate structured output including value propositions, frameworks, and product page copywriting.

Second, \textbf{inquiry response generation}. Given customer inquiries, the system generates multiple versions of response emails (e.g., initial reply, quotation confirmation, cold outreach), with controllable tone, length, and strategy.

Third, \textbf{foreign trade knowledge QA}. The system provides structured answers to professional foreign trade questions, optionally supported by knowledge base references (in private deployments).

Additionally, EvoAgent supports extensibility via API-based skill management and automatic skill synchronization for new users.

\paragraph{Key System Metrics}

In performance testing, we evaluate memory capacity and response latency.

For memory, the system supports up to 420 dialogue turns with 6 compression events without runtime errors. Within 120-turn scenarios, no memory errors are observed across 17 tests. Even at 200 turns without compression, full accuracy is maintained. After compression, while some raw details are lost, core semantics are preserved. The stress-test script of that version covers the turn count, the compression events and errors; it does not include the memory-accuracy checks described here, and it is not part of the current branch.

Compression is triggered when total context exceeds 64K tokens, balancing memory integrity and system stability. We do not conduct strict benchmark comparisons with standalone GPT-5.2 under identical conditions; thus, no direct quantitative comparison is provided.

In our stress test, EvoAgent ran 420 turns with 6 compression events and no errors.

For latency, average response time is approximately 12s in the early stage (the first 40 turns) and stabilizes at ~24s beyond 120 turns. In our deployment, external tool calls add 5–30s latency depending on complexity; this figure is an operating observation and is not produced by that script. The script reports average latency only for the first 40 turns and after turn 120.

\subsubsection{Limitations and Future Work}

\paragraph{What EvoAgent Cannot Do}

The limitations of the evaluated version include:

First, memory constraints: compression leads to irreversible loss of early dialogue details, and long-term memory lacks automatic pruning mechanisms.

Second, user profiling: updates rely on heuristic frequency thresholds and lack deep semantic inference; cross-user knowledge sharing is not supported.

Third, skill execution: only single-skill invocation is supported; no multi-skill orchestration or parallel execution exists.

Fourth, system engineering: lacks cost tracking, monitoring dashboards, and enterprise-grade authentication (e.g., JWT/OAuth).

Fifth, longitudinal evidence: the evaluation compares systems at a point in time and does not measure improvement as state accumulates over sessions; an ablation over the user state (none, profile only, plus memory, plus skills) and a performance curve over session count remain future work.

From a harness engineering perspective, the fundamental limitation is the unidirectional coupling between online execution and offline evolution loops. Evolution is triggered on demand, outside the online loop, preventing real-time adaptation to skill degradation or intent shifts. In other words, EvoAgent supports delayed learning but not real-time adaptation. Addressing this limitation is a key direction for future system evolution. Several of the limitations above have since been partially addressed in later open-source releases; Appendix~C summarizes those changes, which are not evaluated in this paper.

\subsection{Threats to Validity}
\label{sec:threats}

\paragraph{Baseline configuration.} In the scripts used for these experiments, the baseline is a single call to the backbone model at temperature 0 that receives only the final query of each instance, without a system prompt, dialogue context, user profile, memory or skills. The EvoAgent condition runs the full loop on the same query with the case's user and session identifiers, so it can draw on the stored session history, profile, memory and skills. The measured gain is therefore the gain of the harness as a whole, including context that the baseline never sees; it does not isolate the contribution of the evolution mechanism, and we report no ablation over components. The scripts in the current repository add a context-matched baseline that sees the same dialogue, persona, profile and memory but has no tools or skills; the reported numbers do not use it.

\paragraph{Judge.} Responses are scored by an LLM judge (Section~\ref{4.1:RQ1}). In those scripts the judge model followed the same model setting as the system under test. In RQ1 the backbone therefore judges its own outputs, which can favor its own style, and in RQ2 each swapped backbone judges its own pair of responses, so the scores of different backbones are not produced by a common judge and the relative comparisons in Section~\ref{4.2:RQ2} are indicative only. Each pair is scored once with the order of the two responses randomized, which controls position bias but not variance across judge calls; pairs whose judgment cannot be parsed are skipped; and those scripts include no calibration of the judge against human ratings. The scripts in the current repository require the judge to be named and to differ from the model under test, can judge every pair in both presentation orders, and report the pairs that cannot be scored; the reported numbers do not use them.

\paragraph{Statistics.} Each instance is run once per condition, and we report means without confidence intervals or significance tests. We therefore describe differences as substantial rather than significant.

\paragraph{Data.} The 172 evaluation instances are distilled from 664 sessions that were generated by a parameterized script inside a deployed EvoAgent environment, so their distribution is not independent of the system (Section~\ref{4.1.1}). The corpus is proprietary and cannot be released, so the numbers are for reference only; the evaluation scripts and the input schema are released in the repository so that others can evaluate on their own data.

\paragraph{Scope.} The experiments use the evaluated version. The mechanisms of Section~\ref{sec:controlled} and Appendix~C are not evaluated here, and we do not measure how performance changes as state accumulates over sessions.

\section{Conclusion}

This paper proposes EvoAgent, an evolvable unified agent framework designed for real-world foreign trade scenarios. Through structured skill representation, a user-driven self-evolution mechanism, and a hierarchical task delegation architecture, EvoAgent establishes a closed-loop system that integrates skill learning, task execution, and long-term memory management within a harness of reasoning, context engineering and tools, and confines self-improvement to a persistent, user-owned state of profile, memory and skills. 

From a theoretical perspective, we formalize the evolutionary process as a Markov Decision Process (MDP) and introduce an analytical view in which overall model capability is jointly determined by intrinsic inference ability and agent–model synergy. Empirically, EvoAgent raised the judged score of GPT-5.2 in a single run, while GPT-4.1 and Qwen3.5-35B-A3B scored lower with it; this suggests, but does not establish, that the fit between model and harness matters. 

Although the current system still faces limitations in memory compression, multi-skill coordination, and operational support, EvoAgent documents one practical way to build an agent system that accumulates user-specific skills in real-world deployment. Measuring how performance evolves as state accumulates over sessions, and extending the same confirmation-gated loop to the improvement procedure itself, are directions for future work.

\paragraph{Scope of this report.} This paper is an engineering report: it shares practical techniques from building and operating EvoAgent in a foreign trade setting, and the experiments illustrate how these techniques behave in that setting rather than serving as a controlled benchmark. The business data behind the experiments cannot be open-sourced, so the reported results are for reference only; the released code and evaluation scripts document the techniques, not the data.

\section*{Acknowledgment}

This work was conducted as part of an internal industry project at Focus Technology Co., Ltd. 
The authors would like to thank the engineering and product teams for their 
support in system deployment and real-world testing. 
The views expressed in this paper are solely those of the authors and do not 
necessarily represent the official position of the company.

\bibliographystyle{unsrt}  
\bibliography{references}  
\section*{Appendices}
\subsection*{Appendix A: EvoAgent Self-Evolution Process Pseudocode}
\label{appd:a}

\begin{lstlisting}[basicstyle=\ttfamily\footnotesize,frame=single,breaklines=true]
Algorithm : EvoAgent Skill Evolution
% Note: This pseudocode follows the current open-source release (OpenAI Agents SDK).
% Tags: [CE] context engineering, [R] reasoning, [T] tools. Phase-to-stage map (Section 3.3):
% phases 2-3 = stages 1-2; phases 1, 3.5, 4 = stage 3; phase 6 = stages 4-5; phase 7 = stage 6.

Input:
  u           - User input
  S_skills    - Skills Base
  u_profile   - USER.md
  h           - Conversation History
  lambda      - History Compression Threshold (MAX_HISTORY_TOKENS x COMPRESSION_THRESHOLD)
  theta       - Embedding Match Threshold (Default 0.75)

Output:
  s_out       - Matched Skill
  h'          - Updated Conversation History
  u_profile'  - Updated User Profile
  S_skills'   - Updated Skill Base
  c'          - Updated Compression Status

----------------------------------------

1: // Initialization
2: user_id <- extract_user_id(u)
3: session_id <- extract_session_id(u)
4: ws <- Workspace(user_id)
5: h, SOUL, u_profile, memory <- ws.load(session_id)   // SOUL: fixed persona; u_profile, memory: user state; h: agent state

6: // -- Phase 1: History Compression (before skill injection) [CE] --
7: if tokens(h) >= lambda then
8:    assets <- extract_asset_index(h)           // skills, references, images, URLs
9:    summary <- llm_generate_summary(h)         // six-section <summary> block
10:   h <- {summary, assets} + recent_tail(h)    // tool calls and results stay paired
11: end if

12: // -- Phase 2: Skill Matching (Three Stages) [CE] --
13: skill <- NONE
14: // Stage 1: Keyword Matching
15: for each sk in S_skills do
16:    for each t in sk.triggers do
17:        if t.lower() in u.lower() then
18:            skill <- sk
19:            match_type <- "keyword"
20:            confidence <- 1.0
21:            goto PHASE3
22:        end if
23:    end for
24: end for

25: // Stage 2: Embedding Matching (cosine similarity)
26: if skill = NONE then
27:    e_u <- get_embedding(u)                   // Call embedding API
28:    best_score <- 0
29:    for each sk in S_skills do
30:        e_sk <- get_skill_embedding(sk)       // Cached per skill description
31:        score <- cosine_similarity(e_u, e_sk)
32:        if score > best_score then
33:            best_score <- score
34:            skill <- sk
35:        end if
36:    end for
37:    if best_score >= theta then
38:        match_type <- "embedding"
39:        confidence <- best_score
40:    else
41:        skill <- NONE
42:    end if
43: end if

44: // Stage 3: LLM Matching (semantic fallback)
45: if skill = NONE then
46:    skill <- LLM_intent_classify(u, S_skills)
47:    if skill != NONE then
48:        match_type <- "llm"
49:        confidence <- 0.7
50:    end if
51: end if

52: PHASE3:
53: // -- Phase 3: Skill Injection & Context Assembly [CE] --
54: if skill != NONE and skill not in h then
55:    // Inject the skill through a synthetic skill(get) call
56:    h <- h + {role: "assistant", tool_calls: [{
57:             id: "skill_autoload_" + uuid(),
58:             function: {name: "skill",
59:                        arguments: {action: "get", name: skill.name}}
60:           }]}
61:    h <- h + {role: "tool", content: format_skill_content(skill)}

62:    // Increment usage tracking metadata
63:    skill.usage_count <- skill.usage_count + 1
64:    skill.updated_at <- now()
65:    ws.skill_set(skill.name, skill_to_md(skill))
66: end if

67: // -- Phase 3.5: Optional Sub-Agent Handoff (if required by skill) [T] --
68: if skill != NONE and skill.requires_sub_agent then
69:    h <- h + {role: "user", content: u}
70:    // The child runs in its own context and streams its output to the user
71:    receipt <- delegate(skill.sub_agent_name or skill.name, task=u,
72:                        task_id=automatic_task_id(skill.name))
73:    h <- h + receipt                          // the parent keeps only the delivery receipt
74:    goto PHASE5
75: end if

76: // -- Phase 4: Task Execution (Main Agent) [R + T] --
77: instructions <- build_instructions(SOUL, u_profile, memory_index, skill_catalog)
78: main_agent <- Agent(name="EvoAgent",
79:                    instructions=instructions,
80:                    tools=internal_tools + plugins)
81: // The history is compacted again between tool steps when it exceeds lambda
82: result <- Runner.run_streamed(main_agent, input=h, max_turns=MAX_STEPS)
83: h <- h + result.to_input_list()

84: PHASE5:
85: // -- Phase 5: Persist Session (agent state) --
86: ws.history_save(session_id, h)               // also after an interruption, with partial results

87: // -- Phase 6: Offline Evolution (on demand; writes only user state) --
88: // Started by the memory-update action, or by delegating to the session-reviewer sub-agent
89: if review_requested then
90:    review_agent <- Agent(
91:        name="SessionUpdater",
92:        instructions=SESSION_UPDATE + SKILL_EXTRACT,
93:        tools=[update_user_profile, memory, skill]
94:    )
95:    review_result <- Runner.run_streamed(review_agent, input=h, max_turns=10)

96:    // update_user_profile and memory calls write USER.md and MEMORY.md
97:    // (conservative update principle: only explicit, reliable information;
    //  suggestions from evidence repeated across conversations wait for user confirmation)
98:    u_profile <- ws.read_prompt("USER")

99:    // The skill tool cannot write skills: it can only propose them
100:   for each p in review_result.skill_proposals do   // skill(action="propose")
101:       stage(p)                              // validated, stored on the server, shown to the user
102:       if user_confirms(p) and base_unchanged(p) then
103:           apply(p)                          // create | update | patch; previous version kept
104:       end if
105:   end for
106: end if

107: // -- Phase 7: Skill Maturity Assessment (frontend display) --
108: if skill != NONE then
109:    n_uses <- skill.usage_count
110:    sr <- skill.success_rate                  // initialized to 1.0; not updated in this release
111:    if n_uses >= 10 and sr >= 0.85 then
112:        maturity <- "Expert"
113:    else if n_uses >= 4 and sr >= 0.7 then
114:        maturity <- "Established"
115:    else if n_uses >= 1 then
116:        maturity <- "Growing"
117:    else
118:        maturity <- "Emerging"
119:    end if
120: end if

121: return skill, h, u_profile, S_skills, c
\end{lstlisting}
\vspace{1.5em}

\subsection*{Appendix B: EvoAgent Git}
\label{APP:B}
EvoAgent: \url{https://github.com/Focus-AI-Center/Mentarc-EvoAgent}. The scripts used for RQ1--RQ3 are no longer part of the current branch; it keeps two updated scripts in the \texttt{evaluation/} directory, one that generates responses under each condition and one that scores them with an LLM judge; the evaluation corpus contains proprietary business data that cannot be released, so the reported results are for reference only.

\subsection*{Appendix C: Changes Since the Earlier Open-Source Release}
\label{APP:C}
The experiments were run on the evaluated version (Section~\ref{sec:experiments}). Section~\ref{sec:others}, Figure~\ref{fig:architecture} and Appendix~A describe the current open-source release. The main differences from the earlier open-source release (v1.0), whose code we compared with the current one, are listed below; none of them is covered by the experiments above:

\begin{itemize}
    \item \textbf{Cloud workspace.} A configuration switch selects object-store-backed workspaces in place of the local filesystem. Code execution and stdio MCP servers can run in a per-session Docker container that mounts the workspace; skill and memory tools access storage directly, and concurrent updates to memory files and skill text use conditional writes.
    \item \textbf{Confirmed skill changes.} In the earlier open-source release the confirmation of an extracted skill was enforced only by the front end. Now the \texttt{skill} tool offers read actions and \texttt{propose}, so agents cannot write skills at all; a proposal (create, update, or exact-fragment patch) is validated, stored on the server, shown to the user with every file, and applied only on confirmation. An update is refused if the skill changed in the meantime, reference sizes are bounded, and one previous version is kept so that a change or deletion can be undone (see Section~\ref{sec:controlled}).
    \item \textbf{Tool search and MCP.} Tool schemas can be loaded on demand, and MCP tools can be mounted for a request.
    \item \textbf{Delegation.} Sub-agent tasks carry task identifiers, so independent tasks can run in parallel and be revised later; the parent receives a delivery receipt rather than the child's transcript.
    \item \textbf{Context compression.} The summary prompt has six sections (earlier release: nine parts) and covers the complete transcript within token-bounded fragments; an index of recoverable resources is kept, compression can run between tool steps, and skill guides are unloaded together with their tool grants.
    \item \textbf{Runtime contracts.} Typed tool outcomes, question routing, measured provider usage, a Prometheus metrics endpoint, and saving of partial results when a request is interrupted.
\end{itemize}

\end{document}